\documentclass[12pt]{article}

\usepackage[margin=1in]{geometry}
\usepackage{amsmath}
\usepackage{graphicx}
\usepackage{booktabs}
\usepackage{longtable}
\usepackage{array}
\usepackage{multirow}
\usepackage{hyperref}
\usepackage{cite}
\usepackage{float}
\usepackage{caption}
\usepackage{subcaption}
\usepackage{amssymb}
\usepackage{xspace}
\usepackage[title,titletoc]{appendix}

\graphicspath{{./}}

\title{Multi-Objective Reinforcement Learning for Generating Covalent Inhibitor Candidates}

\author{Renee Gil\\
\small Independent Researcher \\
\small \texttt{renee.gil@mailbox.org}}

\date{}

\begin{document}

\maketitle

\begin{abstract}
Rational design of covalent inhibitors requires simultaneously optimizing 
multiple properties, such as binding affinity,
target selectivity, or electrophilic reactivity. This presents  
a multi-objective problem not easily
addressed by screening alone.
Here we present a machine learning pipeline for generating covalent inhibitor candidates
using multi-objective reinforcement learning (RL),
applied to two targets: epidermal growth factor receptor (EGFR) and
acetylcholinesterase (ACHE).
A SMILES-based pretrained LSTM serves as the
generative model, optimized via policy gradient RL with Pareto crowding distance
to balance competing scoring functions including synthetic accessibility,
predicted covalent activity, residue affinity, and an approximated docking score.
The pipeline rediscovers known covalent inhibitors at rates of up to 0.50\% (EGFR)
and 0.74\% (ACHE) in 10,000-structure runs, with candidate structures achieving
warhead-to-residue distances as short as 5.5~\AA\ (EGFR) and 3.2~\AA\ (ACHE)
after further docking-based screening.
More notably, the pipeline spontaneously generates structures bearing warhead motifs
absent from the training data - including allenes, 3-oxo-$\beta$-sultams, and
$\alpha$-methylene-$\beta$-lactones - all of which have independent literature
support as covalent warheads.
These results suggest that RL-guided generation can explore covalent chemical
space beyond its training distribution, and may be useful as a tool for
medicinal chemists working on covalent drug discovery.
\end{abstract}

\newpage
\section{Introduction}

Covalent inhibitors form a permanent or long-lived bond with a nucleophilic residue
in the target protein through a two-step mechanism: initial non-covalent recognition
positions the molecule at the binding site, after which the electrophilic warhead reacts
irreversibly with the target residue.
Compared to non-covalent inhibitors, this confers advantages in potency, selectivity,
and residence time, and enables applications where non-covalent binding is insufficient -
including targets with shallow binding pockets, acquired drug-resistance mutations,
and cases requiring sustained target suppression.
\cite{baillieTargetedCovalentInhibitors2016,
ghoshCovalentInhibitionDrug2019,
jacksonCovalentModifiersChemical2017}
The clinical success of ibrutinib and osimertinib demonstrated that covalent inhibitors
can be rationally designed with high target selectivity, reinvigorating this class of
therapeutics after decades of concern over off-target reactivity.
\cite{akinleyeIbrutinibNovelBTK2013,greigOsimertinibFirstGlobal2016}

Their design, however, presents an additional constraint absent in non-covalent drug discovery:
the reactive warhead must be positioned close to the target residue, requiring
careful integration of binding geometry and electrophilic reactivity.
Conventional virtual screening approaches can assess existing compound libraries but
are inherently limited to explored chemical space.

Generative models can sample novel molecules that are not in any training set,
and reinforcement learning provides a principled framework for steering generation
towards user-defined objectives.
\cite{mazuzMoleculeGenerationUsing2023,
bolcatoValueUsing3D2022,
jeonAutonomousMoleculeGeneration2020}
RL-guided molecular generation has shown promise for general drug design;
its application to covalent inhibitors specifically remains underexplored.

In this work, we develop a multi-objective RL pipeline tailored to covalent inhibitor
generation, incorporating scoring functions specific to covalent drug discovery:
a graph convolutional network covalent activity classifier (GCNII, from \cite{gilGraphNeuralNetworks2024}) and
a graph attention network residue affinity classifier (GAT, developed in this work),
both trained on ProteinReactiveDB \cite{gilGraphNeuralNetworks2024},
alongside a new graph neural network (GNN) docking score approximation, synthetic accessibility,
structural overlap and similarity to known inhibitors, and drug-likeness.
We apply this pipeline to EGFR (targeting Cys797) and ACHE (targeting Ser200),
demonstrate rediscovery of known inhibitors and generation of structures with suitable
warhead-residue geometry, and report the spontaneous emergence of atypical warhead
scaffolds with no precedent in our training data.

\newpage
\section{Methods}

\subsection{Pipeline Overview}

\begin{figure}[H]
    \centering
    \includegraphics[width=0.65\textwidth]{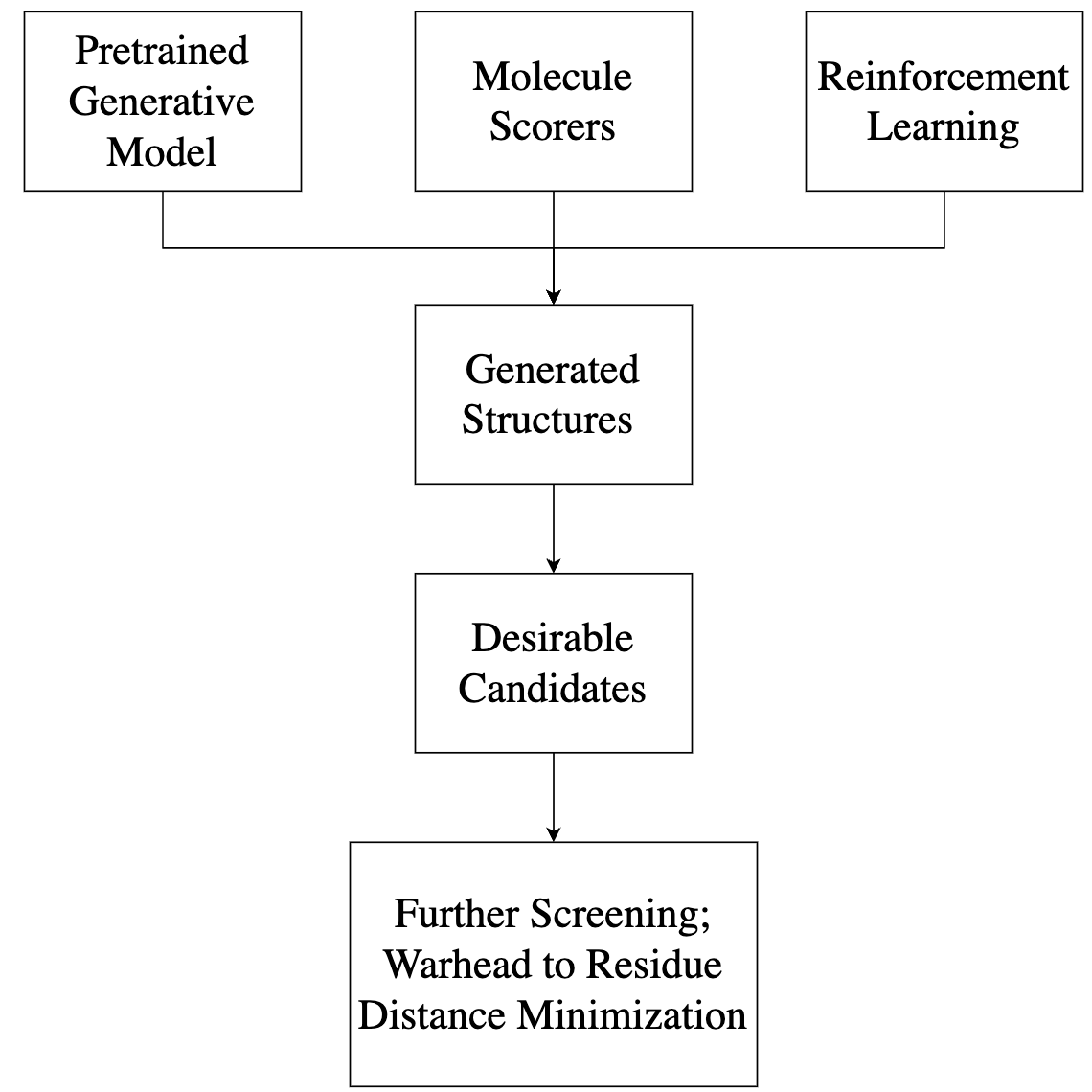}
    \caption{Schematic of the generative RL pipeline.}
    \label{fgr:drug-ex-gen-schema}
\end{figure}

The pipeline consists of three components: a pretrained generative model, a set of
molecular scoring functions, and a reinforcement learning algorithm that optimizes
the generator to maximize the scores.
A schematic is shown in Figure~\ref{fgr:drug-ex-gen-schema}.

\subsection{Generative Model}

Molecule generation used the DrugEx \cite{sichoDrugExDeepLearning2023} framework,
which implements a SMILES-based Long Short-Term Memory (LSTM)
\cite{hochreiterLongShortTermMemory1997} model trained to predict the next character
of a SMILES string.
Rather than training from scratch, we used a publicly available pretrained LSTM
trained on the Papyrus \cite{bequignonPapyrusLargescaleCurated2023} dataset
(1,270,570 unique structures compiled from
ChEMBL \cite{ChEMBL1},
PubChem \cite{10.1093/nar/gkv951},
PDBBind \cite{wangPDBbindDatabaseMethodologies2005},
and BindingDB \cite{liu_bindingdb_2007}),
available in the DrugEx repository
(\url{https://github.com/CDDLeiden/DrugEx},
\texttt{qsprpred} extra).

\subsection{Scoring Functions}

Table~\ref{tbl:drugex-scorers} lists all scoring functions.
Scorers marked with * are included in all models and enforce minimum viability;
the remaining scorers are added incrementally across models (see Section~\ref{sec:models}).

\begin{table}[H]
    \centering
    \caption{Scoring functions used for evaluating generated molecules.
    Scorers marked * are included in all models and enforce minimum viability.
    All raw scores are clipped to $[0,1]$ before being passed to the RL algorithm
    (see Figure~\ref{fgr:drugex-scorer-scaling}); thresholds are applied to the clipped
    scores to define minimally desirable structures.}
    \label{tbl:drugex-scorers}
    \begin{tabular}{p{4.0cm}p{6.8cm}p{3.7cm}}
        \toprule
        \textbf{Scorer} & \textbf{Description} & \textbf{Threshold} \\
        \midrule
        Validity* &
            SMILES grammar validity. &
            must be valid \\
        Synthetic Accessibility (SA)* &
            Synthesizability score of Ertl et al.\
            \cite{ertlEstimationSyntheticAccessibility2009};
            raw scale 1-10 (lower = easier to synthesize). &
            raw $\leq 6$ \\
        Covalent Activity* &
            Probability of covalent activity from the GCNII classifier
            of Gil \& Rowley \cite{gilGraphNeuralNetworks2024}. &
            $p \geq 0.75$ \\
        Covalent Residue Affinity* &
            Probability of reaction with the target residue class
            (Cys or Ser/Thr) from a GAT classifier developed in
            this work (Section~\ref{sec:residue-affinity}). &
            $p \geq 0.75$ \\
        Docking Score Approximation* &
            GNN approximation of AutoDock Vina docking score
            for EGFR or ACHE (Section~\ref{sec:docking-approx});
            raw scores in kcal/mol, more negative = stronger binding. &
            $\leq -6.0$ kcal/mol \\
        Known Structure Overlap &
            Crippen-weighted 3D alignment score
            (RDKit \texttt{GetCrippenO3A} \cite{RdKit})
            against a reference inhibitor. &
            score $\geq 100$ \\
        Tanimoto Similarity &
            Morgan fingerprint similarity ($r=2$, 2048 bits)
            to a reference scaffold. &
            $T \geq 0.1$ \\
        QED \cite{bickertonQuantifyingChemicalBeauty2012} &
            Quantitative drug-likeness estimate based on MW,
            LogP, H-bond counts, and related properties. &
            none \\
        \bottomrule
    \end{tabular}
\end{table}

\begin{figure}[H]
    \centering
    \includegraphics[width=0.85\textwidth]{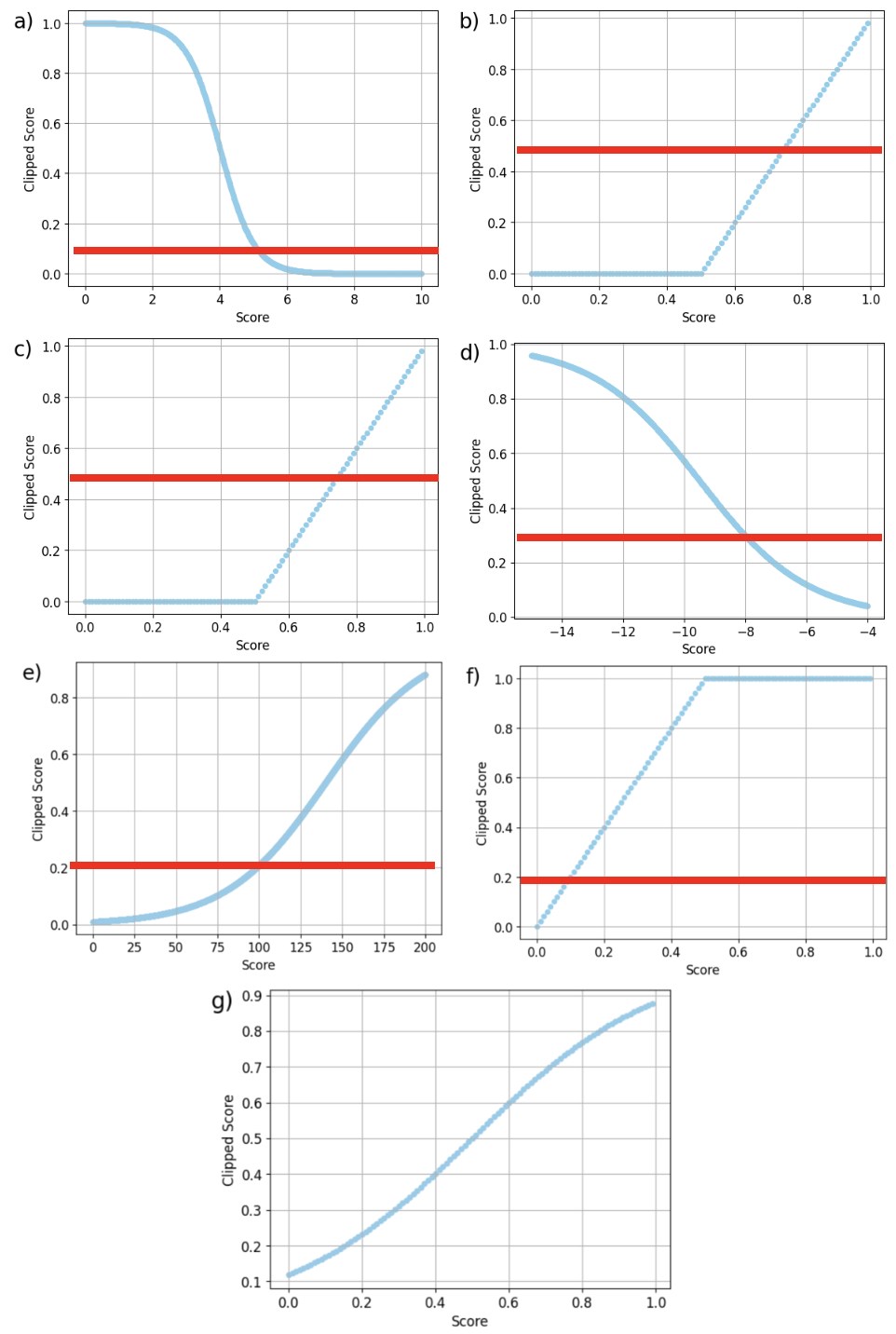}
    \caption{Clipped scoring functions for (a) SA, (b) Covalent Activity,
    (c) Residue Affinity, (d) Docking Score, (e) Overlap,
    (f) Tanimoto Similarity, (g) QED.
    Red lines mark desirability thresholds.
    All clipping functions map the raw score to $[0,1]$; any generated
    structure falling below all active thresholds is counted as desirable.}
    \label{fgr:drugex-scorer-scaling}
\end{figure}

The clipping functions are shown in Figure~\ref{fgr:drugex-scorer-scaling}.

\newpage
\newpage

\paragraph{Threshold selection.}
Thresholds were chosen as follows.
\textbf{SA}: the Ertl et al.\ scale characterizes raw scores above 6 as moderately
to very difficult to synthesize
(1-3 very easy; 3.1-5 moderately easy; 5.1-6.5 moderately difficult; 6.6-10 very
difficult or impractical).
A threshold of 6 was chosen to admit moderately challenging structures while
excluding those likely to be impractical; the clipping function levels off
gradually toward zero rather than applying a hard floor, so the optimizer is
penalized but not blocked for slightly difficult structures.
\textbf{Covalent Activity and Residue Affinity}: both classifiers output a
probability in $[0,1]$.
A threshold of $p \geq 0.75$ was chosen to ensure high confidence that generated
structures are covalently reactive and target the intended residue type.
The clipping function additionally applies a hard floor at $p = 0.5$: any
structure predicted to be more likely non-covalent than covalent receives
zero reward for that scorer.
\textbf{Docking Score Approximation}: AutoDock Vina scores below $-10$ kcal/mol
indicate very strong binding; $-10$ to $-7$ strong; $-7$ to $-5$ moderate;
above $-5$ weak or unlikely.
The threshold of $\leq -6.0$ kcal/mol (moderate-to-strong range) was set
to match the upper tail of the docking score distribution
for EGFR and ACHE structures in ProteinReactiveDB
(Figure~\ref{fgr:docking-distr}), ensuring that only structures predicted
to bind at least moderately well are counted as desirable.
\textbf{Known Structure Overlap}: to calibrate this threshold, we computed
Crippen-weighted 3D alignment scores against osimertinib for four datasets:
the EGFR subset of ProteinReactiveDB, the EGFR subset of BindingDB,
the non-EGFR covalent subset of ProteinReactiveDB, and the non-covalent
subset of ProteinReactiveDB.
EGFR-targeted inhibitors aligned substantially better with osimertinib than
non-EGFR or non-covalent structures; many BindingDB EGFR structures exceeded
scores of 160.
A threshold of 100 was chosen as the point at which EGFR-targeting inhibitors
begin to separate from structurally unrelated compounds.
\textbf{Tanimoto Similarity}: the threshold of $T \geq 0.1$ is intentionally
lenient - low enough to allow diverse scaffold exploration while still
weakly anchoring generation toward quinoline-like scaffolds.
\textbf{QED}: no threshold is applied, because the QED distribution for
known covalent inhibitors is broad and many effective inhibitors have low QED
values; QED is used here as a chemical stability proxy rather than
a drug-likeness gate.

\subsubsection{Covalent Residue Affinity Model}
\label{sec:residue-affinity}

To score the probability that a generated molecule would react with a specific nucleophilic
residue type, we trained a GAT multiclass classifier on
ProteinReactiveDB \cite{gilGraphNeuralNetworks2024}.
Each structure was labelled with its target residue class: Cys, Ser/Thr, Lys/N-terminal,
Asp/Glu, His, or Tyr.
The model was selected by random search with 10-fold cross-validation.
Performance on the validation and external test sets is shown in
Tables~\ref{tab:residue-val} and \ref{tab:residue-test}.
The pipeline uses only the Cys and Ser/Thr outputs, which have the best-supported
performance; the remaining classes are data-limited.

\begin{table}[H]
    \centering
    \caption{GAT residue affinity model - validation set accuracy.}
    \label{tab:residue-val}
    \begin{tabular}{lrr}
        \toprule
        \textbf{Class} & \textbf{Correct / Total} & \textbf{Accuracy (\%)} \\
        \midrule
        Cys         & 230 / 233 & 98.7 \\
        Ser/Thr     & 78 / 89   & 87.6 \\
        Lys/Nt      & 47 / 71   & 66.2 \\
        Asp/Glu     & 4 / 5     & 80.0 \\
        His         & 1 / 3     & 33.3 \\
        Tyr         & 3 / 3     & 100.0 \\
        \bottomrule
    \end{tabular}
\end{table}

\begin{table}[H]
    \centering
    \caption{GAT residue affinity model - external test set accuracy.}
    \label{tab:residue-test}
    \begin{tabular}{lrr}
        \toprule
        \textbf{Class} & \textbf{Correct / Total} & \textbf{Accuracy (\%)} \\
        \midrule
        Cys         & 211 / 255 & 82.7 \\
        Ser/Thr     & 34 / 43   & 79.1 \\
        Lys/Nt      & 9 / 31    & 29.0 \\
        Asp/Glu     & 8 / 10    & 80.0 \\
        His         & 4 / 8     & 50.0 \\
        Tyr         & 1 / 5     & 20.0 \\
        \bottomrule
    \end{tabular}
\end{table}

\subsubsection{Docking Score Approximation}
\label{sec:docking-approx}

\begin{figure}[H]
    \centering
    \includegraphics[width=1.0\textwidth]{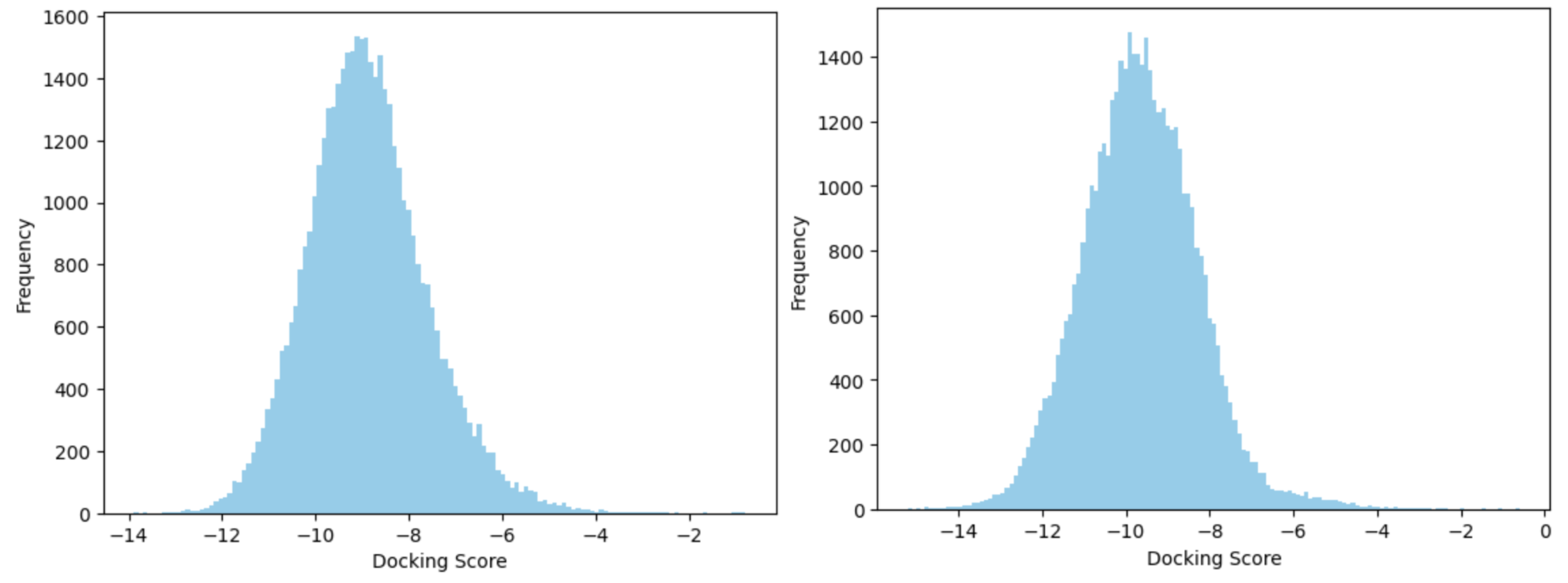}
    \caption{Docking score distributions for EGFR (left) and ACHE (right)
    computed on ProteinReactiveDB.}
    \label{fgr:docking-distr}
\end{figure}

Full docking is impractical at RL scale (tens of thousands of molecules per iteration),
so we trained GNN models to approximate AutoDock Vina scores computed with
DOCKSTRING \cite{garcia-ortegonDOCKSTRINGEasyMolecular2022} (v0.3.4).
Two models were trained: one on EGFR structures from ProteinReactiveDB and one on ACHE
structures, using molgraph \cite{kensert_molgraph_2022} (v0.5.8) with hyperparameters
selected by random search and 10-fold cross-validation.
Docking score distributions for both targets are shown in Figure~\ref{fgr:docking-distr}.

\subsection{Reinforcement Learning}

The generator is optimized using a policy gradient scheme following
Šícho et al.\ \cite{sichoDrugExDeepLearning2023}:

\begin{equation}
    J(\theta) = \sum_{t=1}^{T} \log G(y_t \mid y_{1:t-1}) \times R^*(y_{1:T})
\end{equation}

\noindent where $G$ is the generator parametrized by $\theta$;
$y_t$ is the $t$-th SMILES token in the generated sequence,
with $t$ running from 1 to the total sequence length $T$;
and $R^*(y_{1:T})$ is the reward for the completed sequence,
defined as a weighted sum of the clipped scoring functions
evaluated on the decoded molecule.
All scorer weights were set to 1.0 (equal weighting).
RL training used DrugEx default hyperparameters.

To maintain population diversity, we apply Pareto crowding distance (PCD):
solutions are ranked by Pareto dominance, and within each front the crowding
distance ensures the optimizer favors well-spread solutions:

\begin{equation}
    \text{CD}_i = \sum_{m} \frac{f_m(i{+}1) - f_m(i{-}1)}{f_m^{\max} - f_m^{\min}}
\end{equation}

\noindent where $i$ indexes a solution within a Pareto front;
$m$ indexes an objective (i.e.\ a scoring function);
$f_m(i{+}1)$ and $f_m(i{-}1)$ are the values of objective $m$ for the
neighboring solutions when the front is sorted by objective $m$;
and $f_m^{\max}$, $f_m^{\min}$ are the maximum and minimum values of
objective $m$ within that front.
Solutions with larger $\text{CD}_i$ are more isolated in objective space;
the optimizer preferentially selects these to maintain a well-spread
population across the Pareto front.

\subsection{Models}
\label{sec:models}

Eight models were trained: four for EGFR and four for ACHE.
For each target, the first model uses only base scorers; subsequent models add
structure overlap, QED, and Tanimoto similarity in that order.
The reference structures used for overlap and similarity were osimertinib
(crystal geometry from PDB 4ZAU \cite{yosaatmadjaBindingModeBreakthrough2015})
for EGFR and malathion (docked structure via DOCKSTRING) for ACHE.
Quinoline was used as the Tanimoto similarity reference scaffold.
A summary of models is given in Table~\ref{tbl:drugex-models}.

\begin{table}[H]
    \centering
    \caption{Model configurations. \checkmark indicates the scorer is active.}
    \label{tbl:drugex-models}
    \begin{tabular}{lcccc}
        \toprule
        \textbf{Model} & \textbf{Base Scorers} & \textbf{Overlap} & \textbf{QED} & \textbf{Similarity} \\
        \midrule
        EGFR-1 & \checkmark &  &  &  \\
        EGFR-2 & \checkmark & \checkmark &  &  \\
        EGFR-3 & \checkmark & \checkmark & \checkmark &  \\
        EGFR-4 & \checkmark & \checkmark & \checkmark & \checkmark \\
        \midrule
        ACHE-1 & \checkmark &  &  &  \\
        ACHE-2 & \checkmark & \checkmark &  &  \\
        ACHE-3 & \checkmark & \checkmark & \checkmark &  \\
        ACHE-4 & \checkmark & \checkmark & \checkmark & \checkmark \\
        \bottomrule
    \end{tabular}
\end{table}

\subsection{Evaluation}

Each model generated 10,000 structures and filtered them against desirability thresholds.
The \textbf{rediscovery rate} is defined as:

\begin{equation}
    \text{Rediscovery Rate} = 100 \times
    \frac{\text{Known covalent inhibitors rediscovered}}{\text{Desirable structures generated}}
    \label{eq:rediscovery-rate}
\end{equation}

\noindent where the reference set of known covalent inhibitors is the covalent subset of
ProteinReactiveDB \cite{gilGraphNeuralNetworks2024}, which draws in turn from
CovalentInDB \cite{du_covalentindb_2021};
rediscovered structures are identified by their CovalentInDB identifiers
in Appendix~\ref{app:rediscovered}.
Rediscovery was determined by exact InChI match.

Additionally, from each model's desirable structures, the top 250 by predicted docking score
were docked with DOCKSTRING and the warhead-to-residue distance was approximated.
Target residues were Cys797 (EGFR) and Ser200 (ACHE).

\subsubsection{Warhead Tagging via GradCAM}

Identifying which atoms constitute the warhead in a generated structure is
necessary for distance estimation but impractical to do manually at scale.
We address this using Gradient-weighted Class Activation Mapping (GradCAM),
adapted to graph neural networks following Pope et al.\ \cite{pope_explainability_2019}
and first applied to covalent inhibitor identification in \cite{gilGraphNeuralNetworks2024}.

GradCAM computes class-specific importance weights for feature $k$ at layer $l$
by averaging the gradient of the class score $y^c$ over all $N$ nodes:

\begin{equation}
    \alpha_k^{l,c} = \frac{1}{N} \sum_{n=1}^{N}
    \frac{\partial y^c}{\partial F_{k,n}^{l}}
    \label{eq:gradcam-weights}
\end{equation}

\noindent where $F_{k,n}^{l}$ is the value of feature $k$ at node $n$ in layer $l$.
The per-layer node heatmap is then:

\begin{equation}
    L_{\text{GradCAM}}^{c}[l,\,n] =
    \text{ReLU}\!\left(\sum_{k} \alpha_k^{l,c}\, F_{k,n}^{l}(X, A)\right)
    \label{eq:gradcam-heatmap}
\end{equation}

\noindent where $X$ is the node feature matrix, $A$ is the adjacency matrix,
and $c$ is the covalent-active class.
The ReLU retains only positive contributions to the class score.
This is then averaged across all $L$ layers to give the final per-node score:

\begin{equation}
    L_{\text{GradCAM}}^{c}\text{Avg}[n] =
    \frac{1}{L} \sum_{l=1}^{L} L_{\text{GradCAM}}^{c}[l,\,n]
    \label{eq:gradcam-avg}
\end{equation}

In \cite{gilGraphNeuralNetworks2024}, GradCAM applied to the covalent activity GCNII
classifier showed that atoms selected above a normalized threshold of 0.3 corresponded
to the warhead substructure in 179 of 217 (82.5\%) positively-classified compounds in
the external test set, confirming that the model's decisions are grounded in the
electrophilic region of the molecule.

\begin{figure}[H]
    \centering
    \begin{tabular}{cc}
        \includegraphics[width=0.45\textwidth]{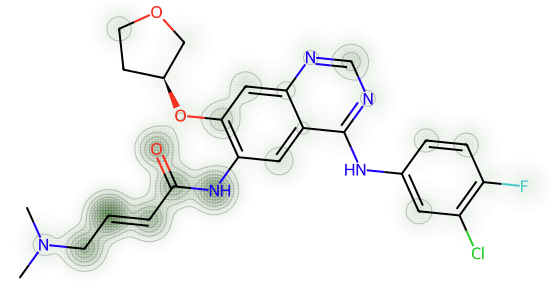} &
        \includegraphics[width=0.45\textwidth]{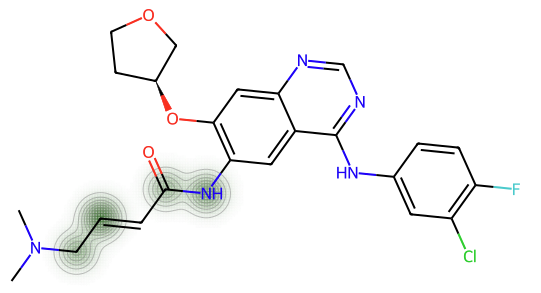} \\
        (a) & (b)
    \end{tabular}
    \caption{Osimertinib GradCAM before (a) and after (b) warhead-focused filtering.
    The full map (a) activates broadly across the molecule; the filtered map (b)
    isolates the acrylamide warhead by retaining only atoms above 0.3 normalized activation.}
    \label{fgr:warhead-tagging}
\end{figure}

For the warhead tagging task here, we retrained the GCNII on a class-balanced dataset
(equal numbers of covalent and non-covalent structures).
This reduces the overall classifier accuracy - the balanced training trades some
covalent-activity prediction performance for sharper localization of the warhead atoms,
since the model is no longer dominated by a majority covalent class and must
discriminate more precisely on local atomic features.
$L_{\text{GradCAM}}^{c}\text{Avg}[n]$ is normalized to $[0,1]$ and atoms above 0.3
are retained as the warhead region.
Figure~\ref{fgr:warhead-tagging} shows the effect of this filtering on osimertinib:
the full GradCAM map highlights much of the molecule, while the filtered map isolates
the acrylamide warhead.

Chemical space coverage was visualized using Morgan fingerprints (radius 2, 2048 bits)
followed by PCA (50 components) and t-SNE (perplexity 30, 1000 iterations).

\section{Results}

\subsection{Docking Score Approximation Models}

\begin{figure}[H]
    \centering
    \begin{subfigure}[b]{0.48\textwidth}
        \includegraphics[width=\textwidth]{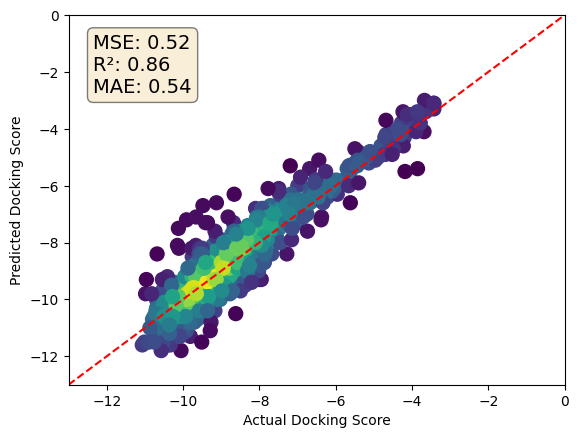}
        \caption{EGFR docking score approximation.}
        \label{fgr:egfr-docking}
    \end{subfigure}
    \hfill
    \begin{subfigure}[b]{0.48\textwidth}
        \includegraphics[width=\textwidth]{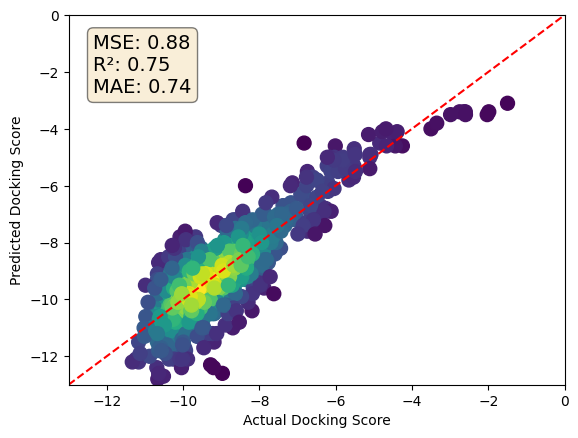}
        \caption{ACHE docking score approximation.}
        \label{fgr:ache-docking}
    \end{subfigure}
    \caption{Density scatter plots of predicted vs.\ actual docking scores.}
\end{figure}

The EGFR docking approximation model achieved $R^2 = 0.86$, MSE $= 0.52$, MAE $= 0.54$.
The ACHE model achieved $R^2 = 0.75$, MSE $= 0.88$, MAE $= 0.74$.
Parity plots are shown in Figures~\ref{fgr:egfr-docking} and \ref{fgr:ache-docking}.
Accuracy is consistent across small, medium, large, and very large molecules
(Appendix~\ref{app:docking-size}).

\subsection{EGFR Models}

Table~\ref{tbl:egfr-models-performance} summarizes EGFR model performance.
The best rediscovery rate was achieved by EGFR-3 (0.50\%).
Adding more scoring objectives reduces the number of desirable structures generated,
consistent with the optimizer having a harder time satisfying all constraints simultaneously.
The notably low rate for EGFR-2 (0.18\%) reflects the strong pull of the overlap
scorer toward osimertinib-like structures; adding QED and similarity (EGFR-3, EGFR-4)
restores diversity.

\begin{table}[H]
    \centering
    \caption{EGFR model performance (10,000 structures generated per model).}
    \label{tbl:egfr-models-performance}
    \begin{tabular}{lccc}
        \toprule
        \textbf{Model} & \textbf{Desirable Structures} & \textbf{Rediscovered} & \textbf{Rate (\%)} \\
        \midrule
        EGFR-1 & 7391 & 36 & 0.49 \\
        EGFR-2 & 6776 & 12 & 0.18 \\
        EGFR-3 & 4793 & 24 & 0.50 \\
        EGFR-4 & 4334 & 17 & 0.39 \\
        \bottomrule
    \end{tabular}
\end{table}

\begin{figure}[H]
    \centering
    \includegraphics[trim=0 0 300 0, clip, width=0.75\textwidth]{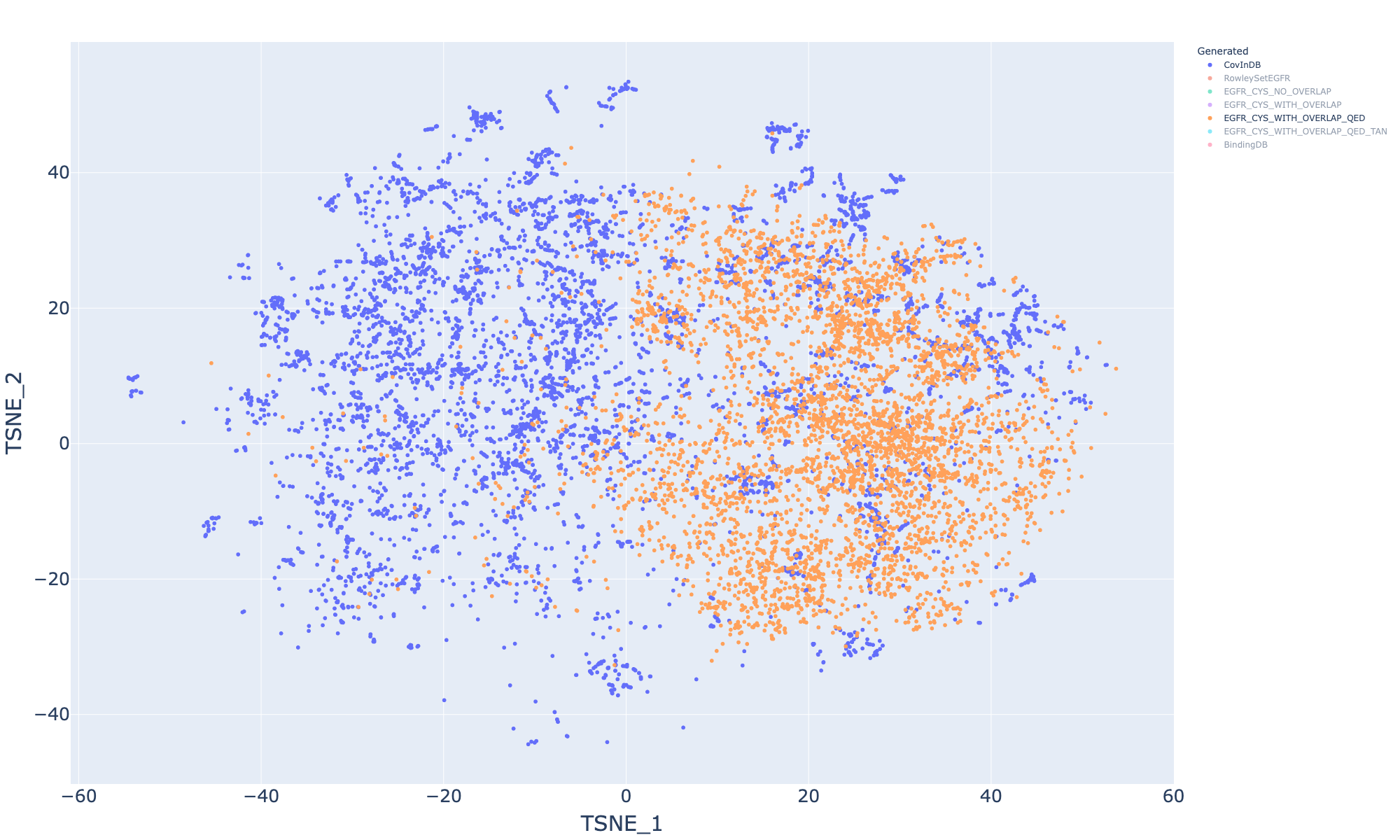}
    \caption{t-SNE embedding of desirable structures from EGFR-3 (orange)
    overlaid on the covalent subset of ProteinReactiveDB (blue).}
    \label{fgr:egfr-3}
\end{figure}

A t-SNE visualization for EGFR-3 (best rediscovery rate) is shown in Figure~\ref{fgr:egfr-3};
all four models occupy a similar region of chemical space, with acrylamides dominating -
expected given that osimertinib is an acrylamide overlap reference on an acrylamide-rich
training set (remaining models in Appendix~\ref{app:tsne}).
Example rediscovered inhibitors are shown in Appendix~\ref{app:rediscovered}.

\subsubsection{Post-Screening: Warhead-Residue Distances}

\begin{figure}[H]
    \centering
    \includegraphics[width=0.6\textwidth]{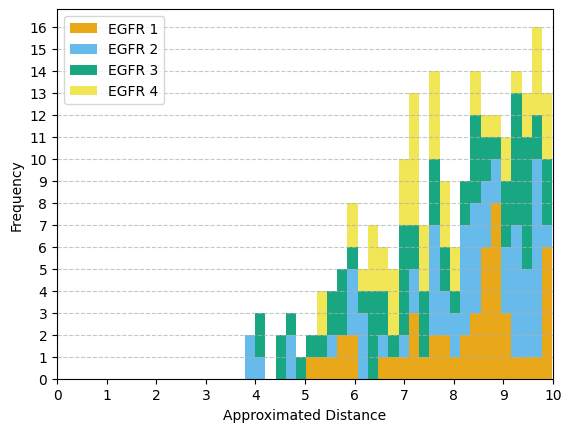}
    \caption{Warhead-to-Cys797 distance distributions for top EGFR candidates.}
    \label{fig:egfr-dist}
\end{figure}

\begin{figure}[h]
    \centering
    \includegraphics[width=0.45\textwidth]{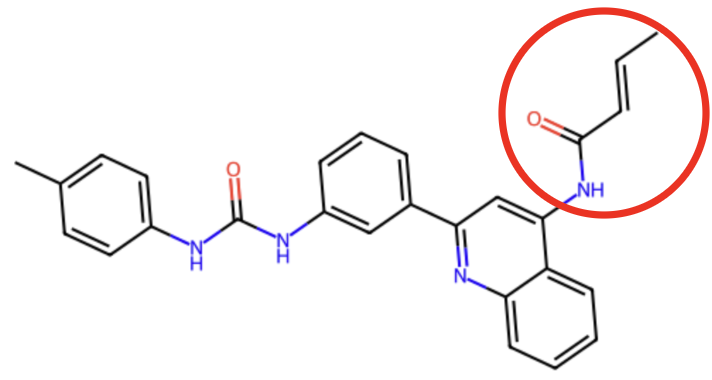}
    \includegraphics[width=0.60\textwidth]{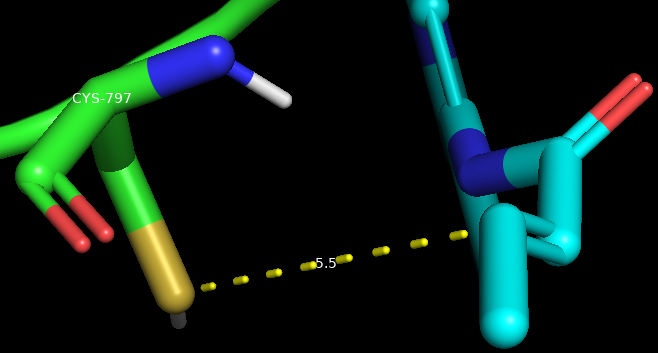}
    \caption{Best EGFR candidate: structure with warhead circled (top) and
    docking distance visualization (bottom). Warhead-to-Cys797 distance is 5.5~\AA.}
    \label{fig:egfr-combined}
\end{figure}

Among the top-250 docked structures per model, we filtered for warhead-to-Cys797 distances
below 10~\AA.
Results are shown in Figure~\ref{fig:egfr-dist}.
The shortest approximated distance was 3.78~\AA\ (EGFR-2); full docking of that candidate
confirmed a Cys797-to-warhead distance of 5.5~\AA\ (Figure~\ref{fig:egfr-combined}),
comparable to the 3.35~\AA\ seen in the osimertinib crystal structure (PDB 4ZAU).

\subsubsection{Effect of Generation Volume}

Running EGFR-3 at increasing generation scales
(Table~\ref{tbl:more-generated-structures-egfr}, Appendix~\ref{app:genvolume})
shows that absolute rediscoveries increase monotonically;
the rediscovery rate fluctuates due to the stochastic nature of the generator.

\newpage
\subsection{ACHE Models}

Table~\ref{tbl:ache-models-performance} summarizes ACHE results.
ACHE-3 achieved the best rediscovery rate (0.74\%).
The t-SNE plots (Figures~\ref{fgr:ache-1} and~\ref{fgr:ache-3};
ACHE-2 and ACHE-4 in Appendix~\ref{app:tsne}) reveal a qualitative
shift across models: ACHE-1 generates predominantly boron-containing compounds,
while ACHE-2 through ACHE-4 show a stronger phosphorus presence, driven by the
malathion overlap reference.
Example rediscovered inhibitors are shown in Appendix~\ref{app:rediscovered}.

\begin{table}[H]
    \centering
    \caption{ACHE model performance (10,000 structures generated per model).}
    \label{tbl:ache-models-performance}
    \begin{tabular}{lccc}
        \toprule
        \textbf{Model} & \textbf{Desirable Structures} & \textbf{Rediscovered} & \textbf{Rate (\%)} \\
        \midrule
        ACHE-1 & 9053 & 6  & 0.07 \\
        ACHE-2 & 1835 & 0  & 0.00 \\
        ACHE-3 & 945  & 7  & 0.74 \\
        ACHE-4 & 468  & 0  & 0.00 \\
        \bottomrule
    \end{tabular}
\end{table}

\begin{figure}[H]
    \centering
    \begin{subfigure}[b]{0.48\textwidth}
        \includegraphics[trim=0 0 300 0, clip, width=\textwidth]{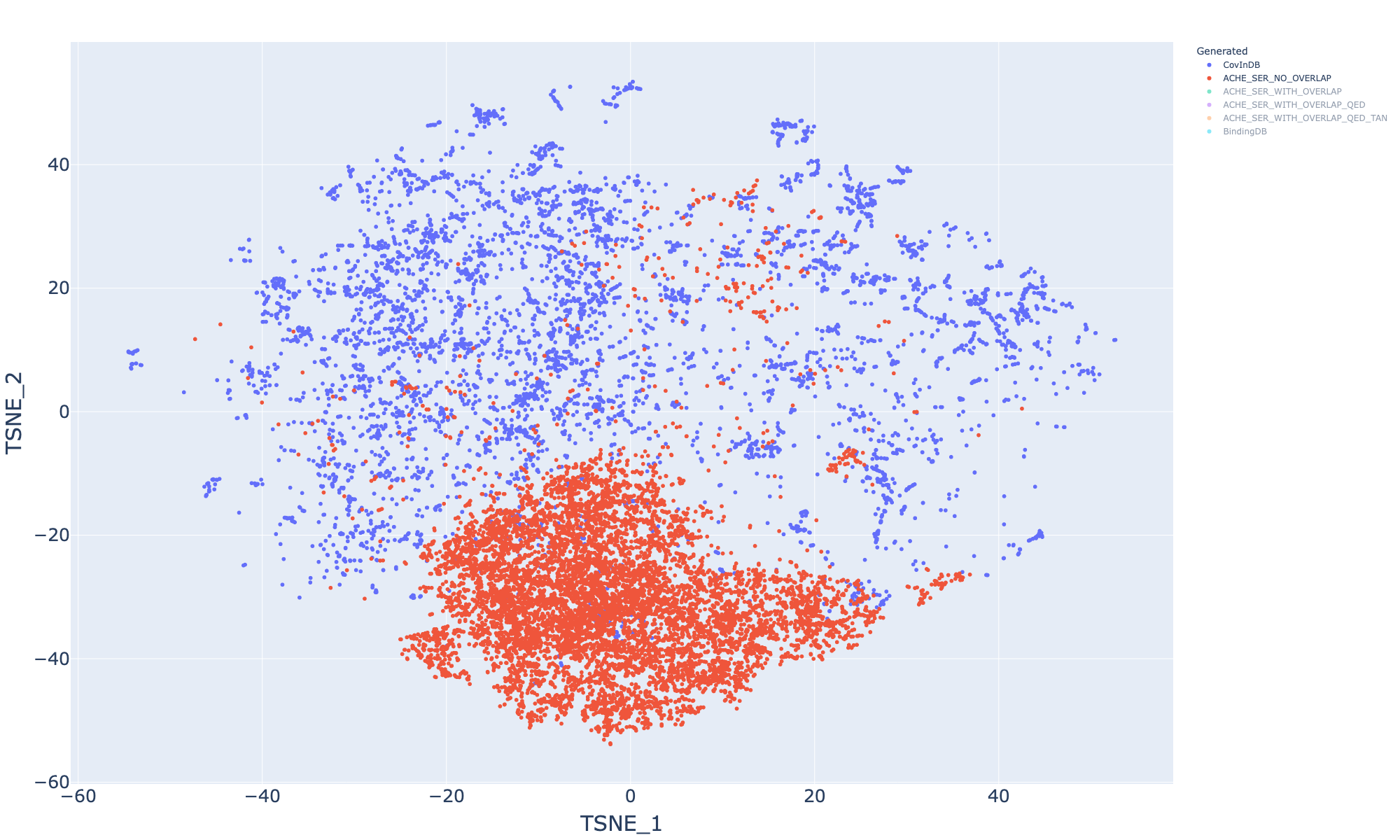}
        \caption{ACHE-1 (orange): boron-dominated space.}
        \label{fgr:ache-1}
    \end{subfigure}
    \hfill
    \begin{subfigure}[b]{0.48\textwidth}
        \includegraphics[trim=0 0 300 0, clip, width=\textwidth]{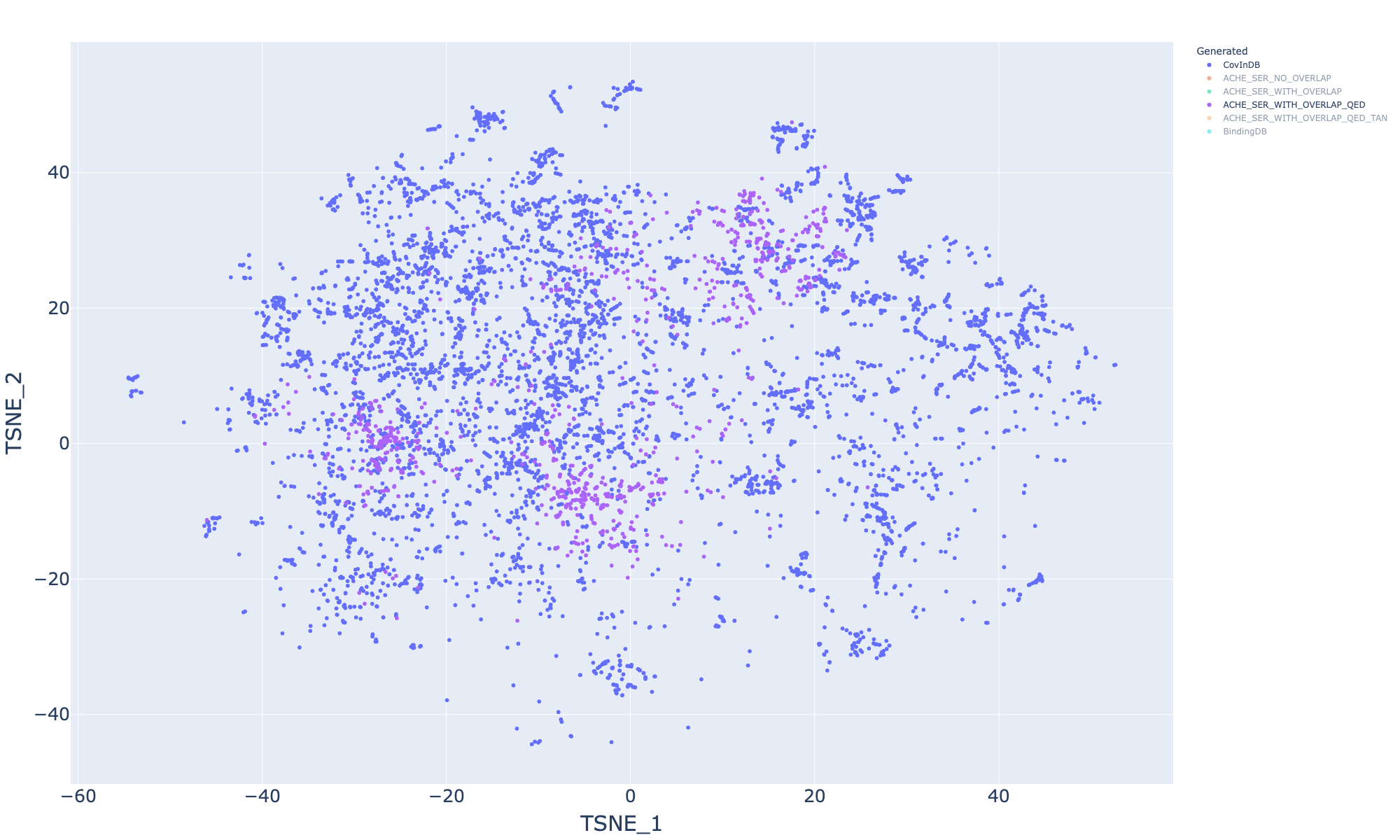}
        \caption{ACHE-3 (purple): phosphorus-enriched space.}
        \label{fgr:ache-3}
    \end{subfigure}
    \caption{t-SNE embeddings of desirable structures from ACHE-1 and ACHE-3 (colored)
    overlaid on the covalent subset of ProteinReactiveDB (blue).
    ACHE-2 and ACHE-4 are shown in Appendix~\ref{app:tsne}.}
\end{figure}

\subsubsection{Post-Screening: Warhead-Residue Distances}

\begin{figure}[H]
    \centering
    \includegraphics[width=0.6\textwidth]{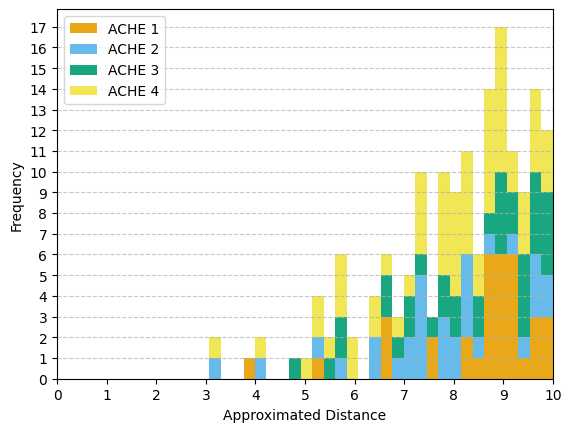}
    \caption{Warhead-to-Ser200 distance distributions for top ACHE candidates.}
    \label{fgr:ache-dist}
\end{figure}

\begin{figure}[H]
    \centering
    \includegraphics[width=0.35\textwidth]{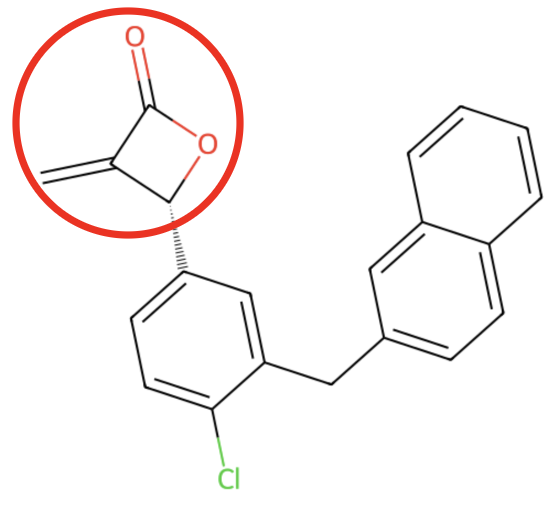}
    \includegraphics[width=0.65\textwidth]{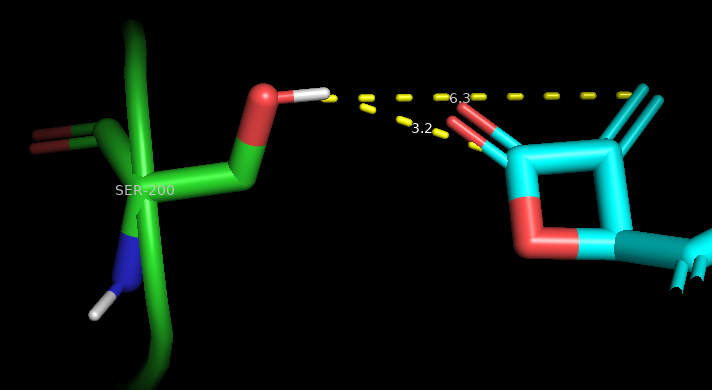}
    \caption{Best ACHE candidate: structure with warhead circled (top) and docking
    visualization (bottom). The warhead is an exocyclic $\alpha$-methylene-$\beta$-lactone;
    distances measured to both the carbonyl carbon (3.2~\AA) and exocyclic double bond (6.3~\AA).}
    \label{fig:ache-combined}
\end{figure}

The shortest approximated warhead-to-Ser200 distance was 3.23~\AA\ (ACHE-4);
full docking of that candidate confirmed distances of 3.2~\AA\ (carbonyl path)
and 6.3~\AA\ (Michael acceptor path),
as shown in Figure~\ref{fig:ache-combined}.
Distance distributions are shown in Figure~\ref{fgr:ache-dist}.

The ACHE best candidate is itself notable: an exocyclic $\alpha$-methylene-$\beta$-lactone.
This motif is not present in ProteinReactiveDB, yet Wang et al.\ have demonstrated that
$\alpha$-methylene-$\beta$-lactones act as dual electrophilic warheads capable of reacting
with Cys, Ser, Thr, Tyr, and Lys residues. \cite{wangAMethylenevLactoneScaffoldDeveloping2021}
This finding motivated a broader search for atypical warheads in the generated structures.

\subsubsection{Effect of Generation Volume}

ACHE-3 rediscovery rates at increasing generation scales are given in
Table~\ref{tbl:generated-structures-ache} (Appendix~\ref{app:genvolume}).
Notably, the ACHE-3 rediscoveries at larger scales include inhibitors of serine hydrolases
and proteases other than ACHE:
butyrylcholinesterase (BCHE), monoacylglycerol lipase (MGLL), fatty acid amide hydrolase (FAAH),
and thrombin - all of which share a reactive serine residue.
This indicates partial drift away from ACHE specifically towards serine-reactive space more broadly,
which is mechanistically sensible but underscores the difficulty of encoding full protein-level
specificity through scoring functions alone.

\subsection{Atypical Warhead Scaffolds}

\begin{figure}[H]
    \centering
    \includegraphics[width=0.5\textwidth]{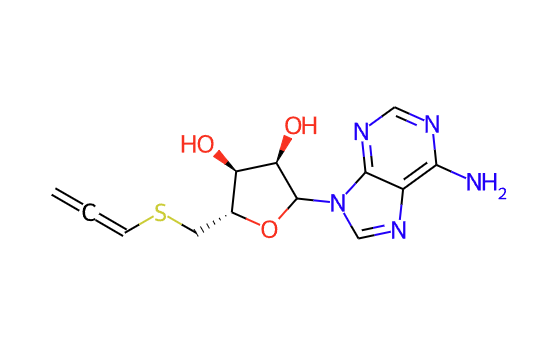}
    \caption{The single allene-containing structure found in ProteinReactiveDB:
    an S-allenyl covalent inhibitor of S-adenosyl-\textsc{l}-homocysteine hydrolase.}
    \label{fig:allene-hit}
\end{figure}

We searched the desirable structures from all models for warhead motifs absent from
ProteinReactiveDB by visual inspection of generated structures, followed by
substructure search of ProteinReactiveDB to confirm absence.
Table~\ref{tbl:untypical-structures} shows three classes of atypical scaffolds found.
A substructure search of ProteinReactiveDB returned zero hits for
3-oxo-$\beta$-sultam and $\alpha$-methylene-$\beta$-lactone, and a single hit for
allene - an S-allenyl inhibitor of S-adenosyl-\textsc{l}-homocysteine hydrolase
\cite{guillermSynthesisMechanismAction2001} (Figure~\ref{fig:allene-hit}),
which targets a different enzyme and bears little structural resemblance to
the allene-containing structures generated here.

These scaffolds therefore appear to be genuine exploration products of the RL pipeline
rather than artifacts of the training data:

\begin{itemize}
    \item \textbf{Allenes}: Tasdan et al.\ have reported an allene-based warhead targeting
    a histidine residue in \textit{E. coli} DsbA. \cite{tasdanIdentificationAlleneWarhead2024}
    \item \textbf{3-Oxo-$\beta$-sultams}: Reported as sulfonylating electrophiles
    that target serine residues. \cite{carvalho3OxovsultamSulfonylatingChemotype2020,
    tsangReactivitySelectivityInhibition2007}
    \item \textbf{$\alpha$-Methylene-$\beta$-lactones}: Dual electrophilic warheads
    reacting with multiple nucleophilic residue types. \cite{wangAMethylenevLactoneScaffoldDeveloping2021}
\end{itemize}

\begin{longtable}{|c|c|}
    \caption{Examples of atypical warhead structures generated by the pipeline.}
    \label{tbl:untypical-structures} \\
    \hline
    \textbf{Model} & \textbf{Structure} \\
    \hline
    \endfirsthead
    \multicolumn{2}{c}{{\bfseries \tablename\ \thetable{} - continued}} \\
    \hline
    \textbf{Model} & \textbf{Structure} \\ \hline
    \endhead
    \hline \multicolumn{2}{r}{{Continued on next page}} \\ \endfoot
    \hline \endlastfoot
    EGFR-1 (allene) & \includegraphics[width=0.48\textwidth]{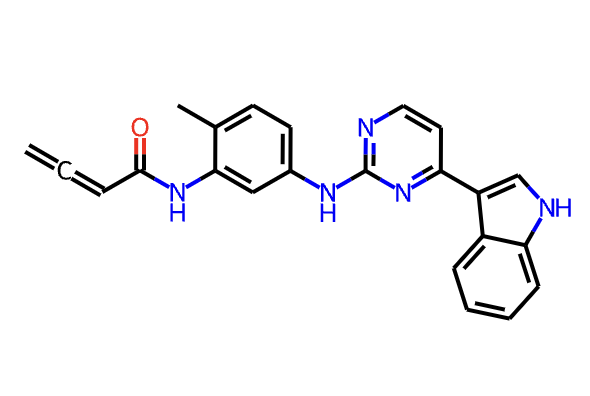} \\
    \hline
    EGFR-4 (allene + macrocycle) & \includegraphics[width=0.40\textwidth]{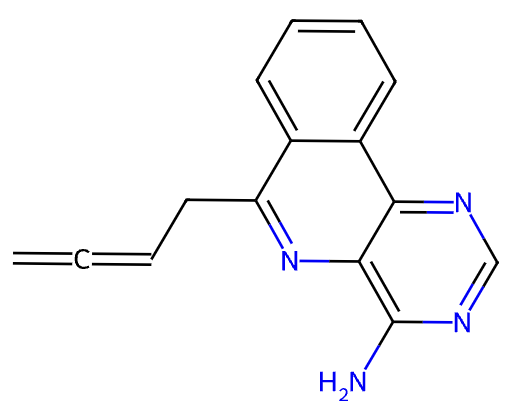} \\
    \hline
    ACHE-2 (3-oxo-$\beta$-sultam) & \includegraphics[width=0.33\textwidth]{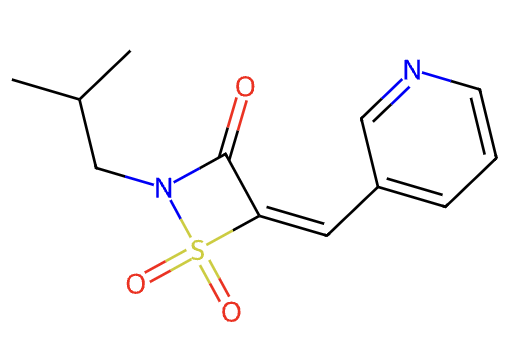} \\
\end{longtable}

\section{Limitations}

The pipeline's scoring functions are themselves ML models trained on limited data.
Because the RL objective is defined entirely by these scores, the generator is
constrained to a chemical space implicitly encoded in the training sets -
including their biases.
Adding more scoring functions narrows the feasible space and can cause the optimizer
to fail to satisfy all constraints simultaneously, as seen in the low desirable-structure
counts for EGFR-4 and ACHE-4.

Several scoring functions are approximations (docking score, warhead tagging),
and their errors compound.
The warhead tagging model tags a region rather than a specific atom,
so reported distances are estimates; the actual reactive geometry must be confirmed
by full docking or ideally by experimental validation.

Finally, as with any computational pipeline, experimental confirmation is the only
definitive test of inhibitory activity.

\section{Conclusions}

We have presented a multi-objective RL pipeline for generating covalent inhibitor
candidates, combining a Papyrus-pretrained LSTM generator with scoring functions
specific to covalent drug discovery.
Applied to EGFR and ACHE, the pipeline generates structures that overlap with
known covalent inhibitor chemical space and rediscovers known inhibitors at rates
of up to 0.50\% (EGFR) and 0.74\% (ACHE).
Further docking-based screening yields candidates with warhead-to-residue geometries
consistent with covalent binding (5.5~\AA\ for EGFR, 3.2~\AA\ for ACHE).

The most striking result is the spontaneous generation of three classes of atypical
warhead scaffolds - allenes, 3-oxo-$\beta$-sultams, and $\alpha$-methylene-$\beta$-lactones -
none of which are present in ProteinReactiveDB yet all of which have independent
literature support as covalent warheads.
This suggests that RL-guided generation can meaningfully extrapolate beyond its
training distribution in chemically relevant directions.

We anticipate this pipeline could serve as a foundation for covalent inhibitor
discovery campaigns, particularly when complemented by experimental screening
of top-ranked candidates.

\section*{Acknowledgments}

This work was conducted as part of the author's doctoral research at Carleton University, Canada.
The author thanks Prof. Christopher Rowley for his supervision and for permission to publish this
work independently.

\subsection*{AI Disclosure}

Claude Sonnet 4.5 (Anthropic) was used for \LaTeX{} debugging and minor text editing during the preparation
of this manuscript. All scientific content, methodology, results, and conclusions are the author's own.

\newpage
\begin{appendices}

\section{Docking Approximation Accuracy by Molecule Size}
\label{app:docking-size}

\begin{figure}[H]
    \centering
    \begin{subfigure}[b]{0.48\textwidth}
        \includegraphics[width=\textwidth]{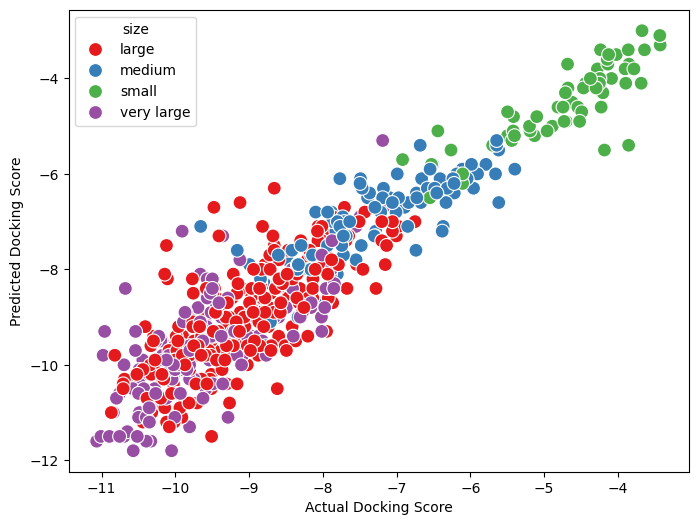}
        \caption{EGFR, by molecule size.}
        \label{fgr:egfr-docking-size}
    \end{subfigure}
    \hfill
    \begin{subfigure}[b]{0.48\textwidth}
        \includegraphics[width=\textwidth]{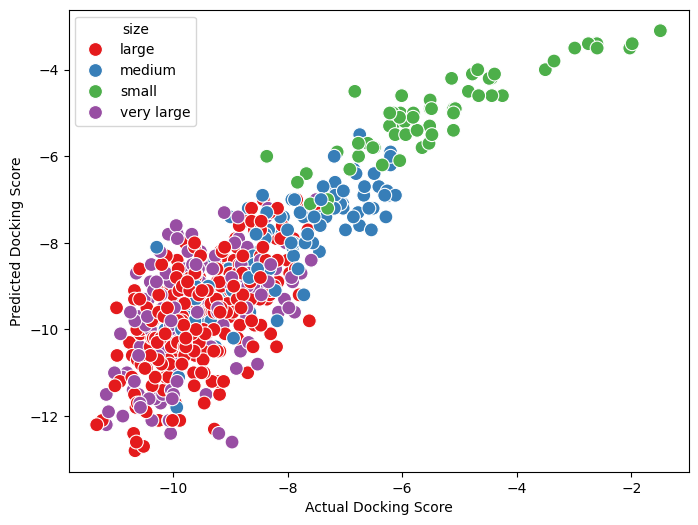}
        \caption{ACHE, by molecule size.}
        \label{fgr:ache-docking-size}
    \end{subfigure}
    \caption{Docking approximation accuracy stratified by molecular size
    (small: $\leq 10$ heavy atoms; medium: 10-20; large: 20-35; very large: $>35$).
    Performance is consistent across all size bins.}
\end{figure}

\section{t-SNE Embeddings}
\label{app:tsne}

\begin{figure}[H]
    \centering
    \begin{subfigure}[b]{0.48\textwidth}
        \includegraphics[trim=0 0 300 0, clip, width=\textwidth]{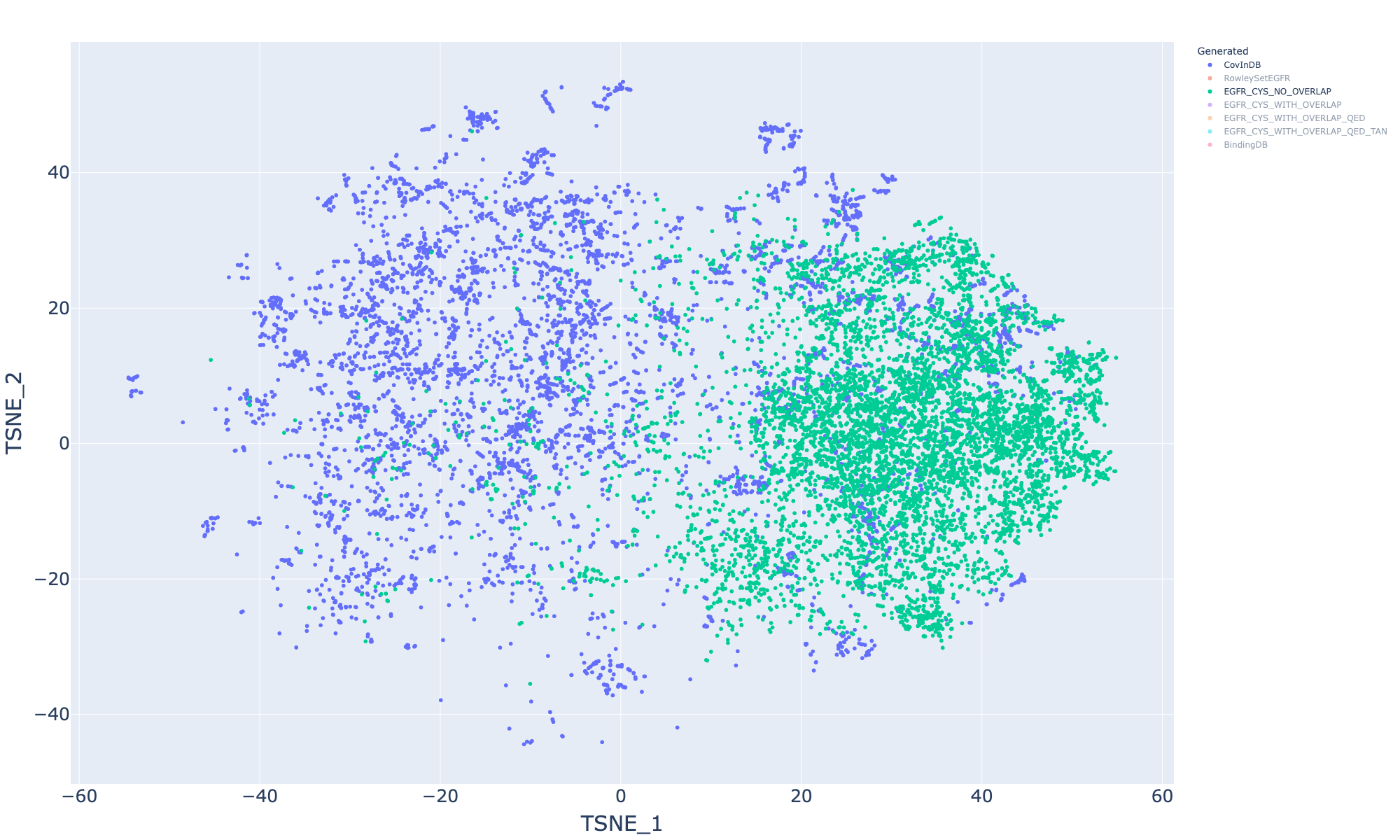}
        \caption{EGFR-1 (green).}
        \label{fgr:egfr-1}
    \end{subfigure}
    \hfill
    \begin{subfigure}[b]{0.48\textwidth}
        \includegraphics[trim=0 0 300 0, clip, width=\textwidth]{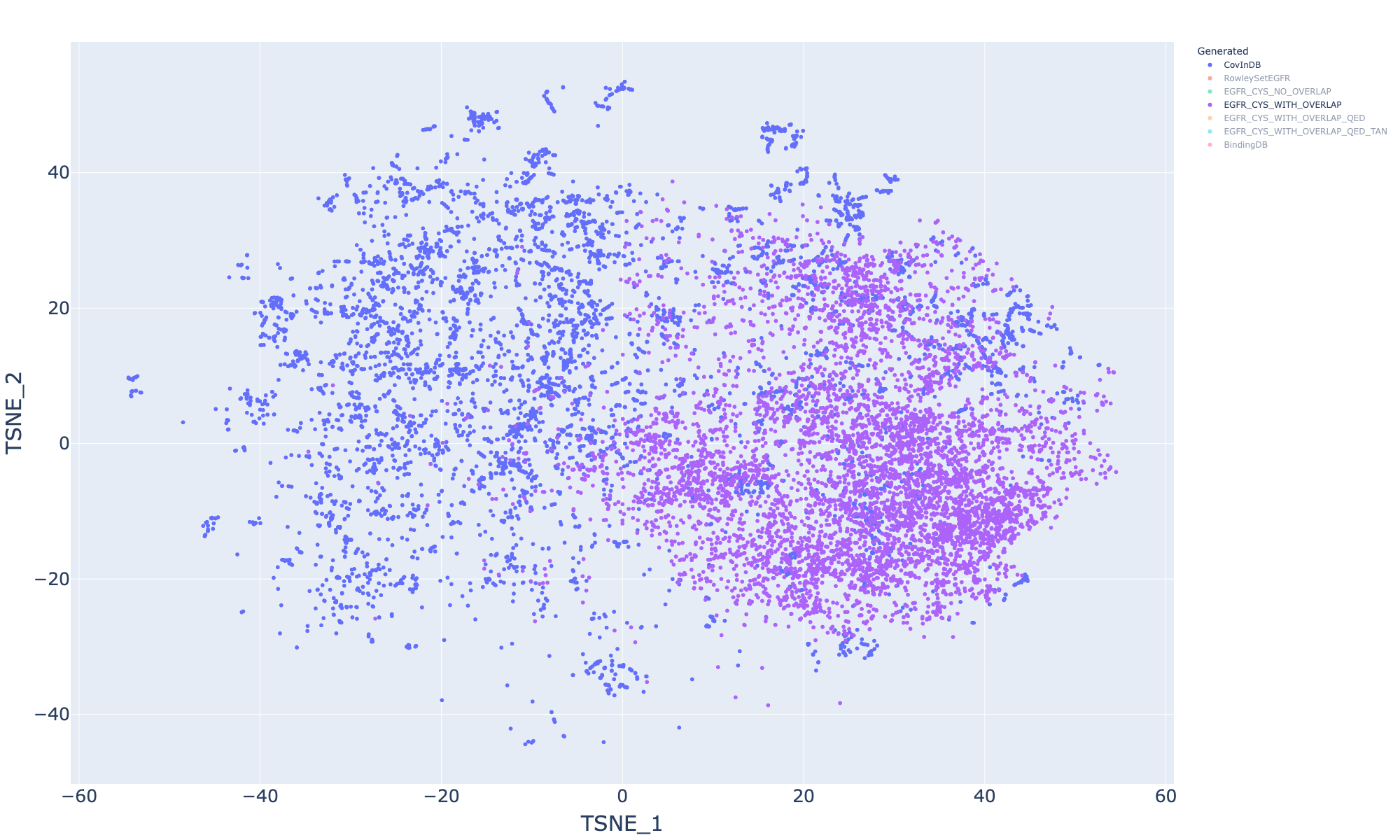}
        \caption{EGFR-2 (purple).}
        \label{fgr:egfr-2}
    \end{subfigure}

    \begin{subfigure}[b]{0.48\textwidth}
        \includegraphics[trim=0 0 300 0, clip, width=\textwidth]{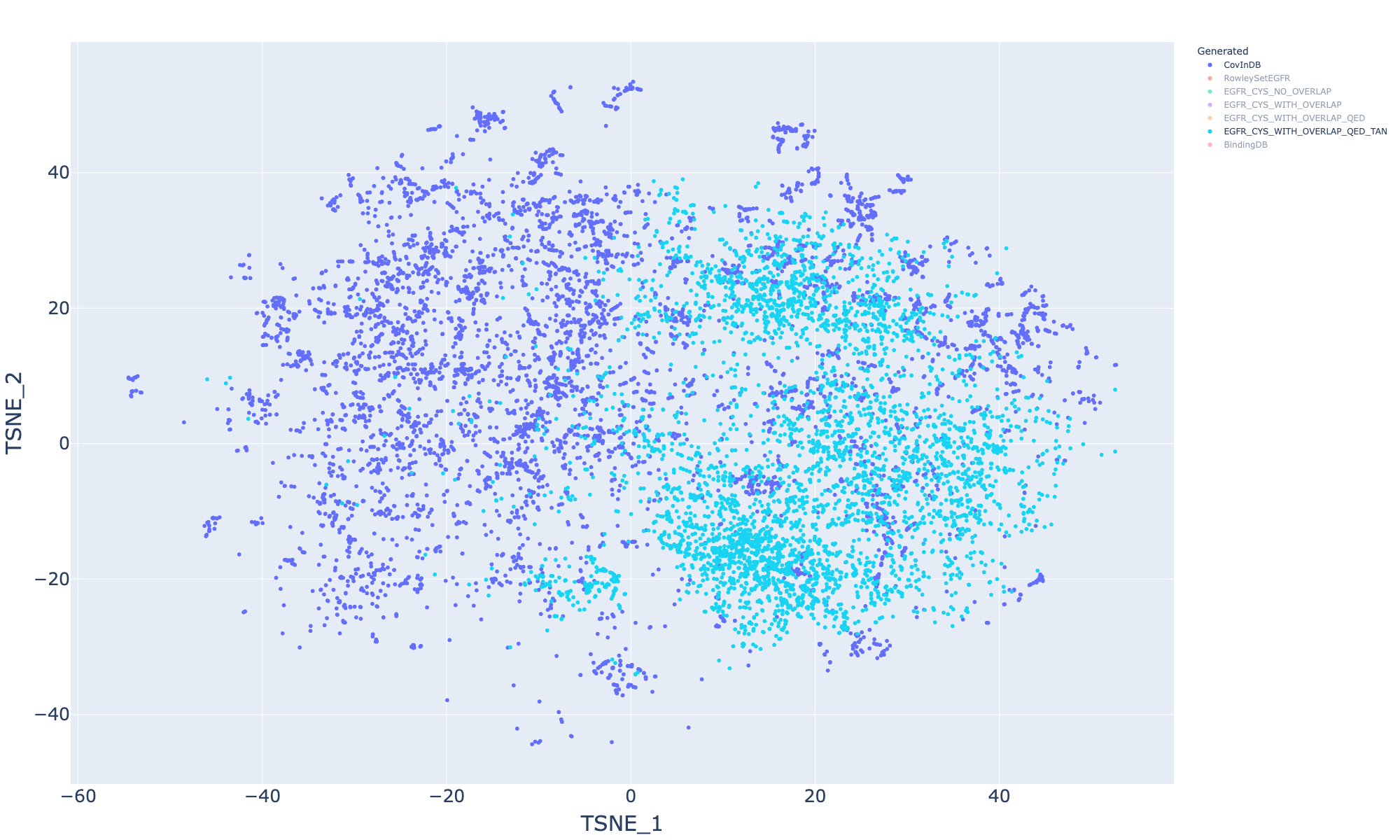}
        \caption{EGFR-4 (teal).}
        \label{fgr:egfr-4}
    \end{subfigure}
    \caption{t-SNE embeddings for remaining EGFR models, overlaid on the
    covalent subset of ProteinReactiveDB (blue). EGFR-3 is shown in the main text
    (Figure~\ref{fgr:egfr-3}). All four models occupy similar chemical space.}
\end{figure}

\begin{figure}[H]
    \centering
    \begin{subfigure}[b]{0.48\textwidth}
        \includegraphics[trim=0 0 300 0, clip, width=\textwidth]{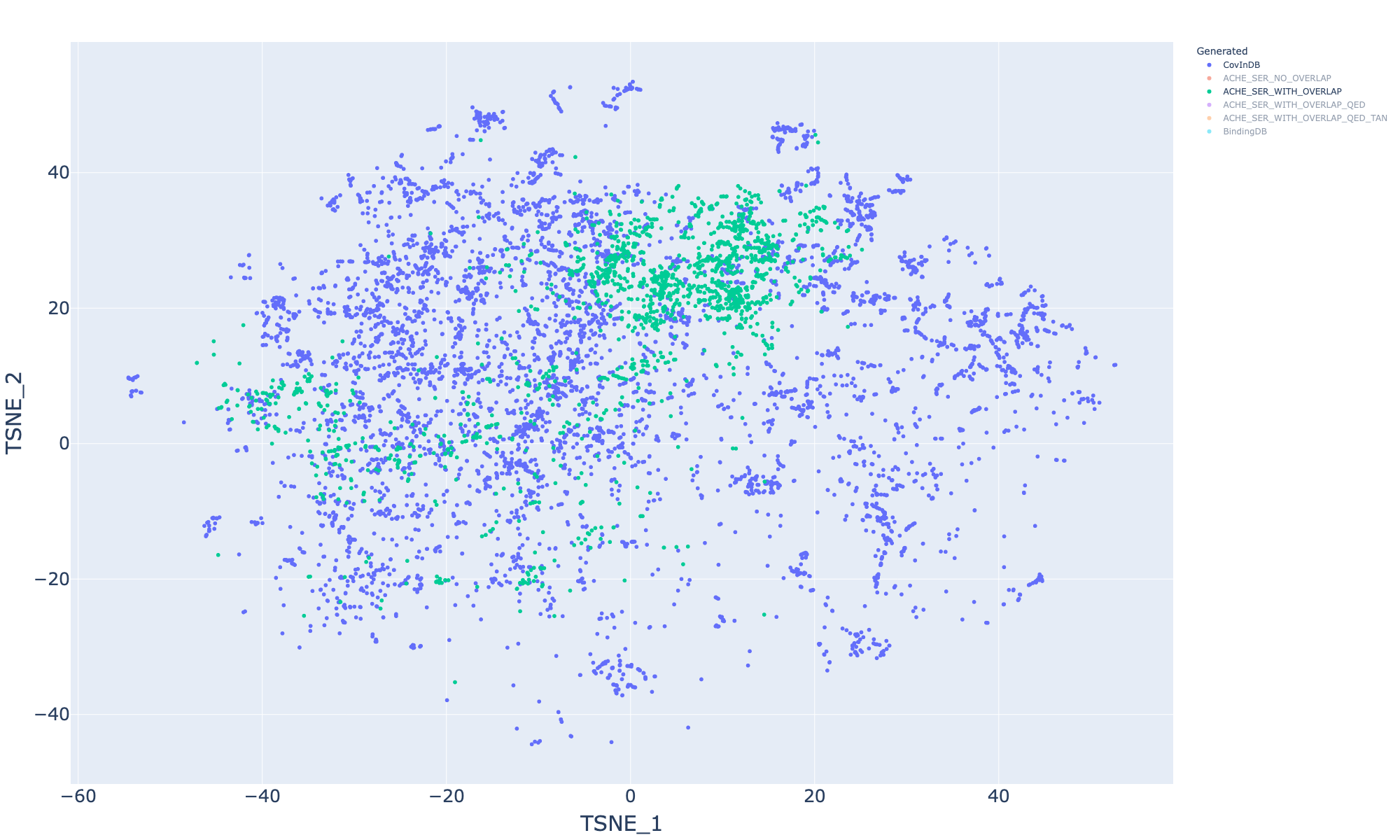}
        \caption{ACHE-2 (teal).}
        \label{fgr:ache-2}
    \end{subfigure}
    \hfill
    \begin{subfigure}[b]{0.48\textwidth}
        \includegraphics[trim=0 0 300 0, clip, width=\textwidth]{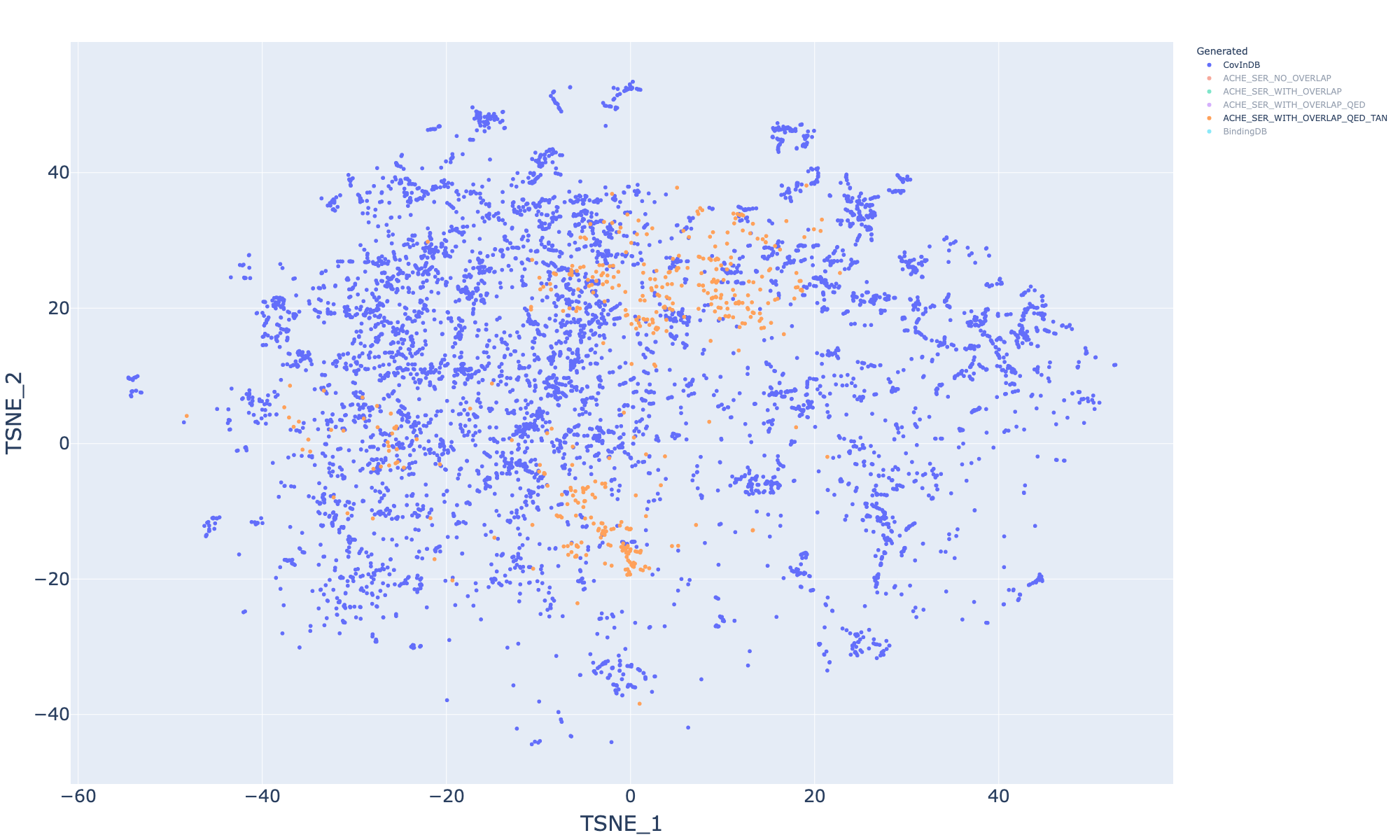}
        \caption{ACHE-4 (orange).}
        \label{fgr:ache-4}
    \end{subfigure}
    \caption{t-SNE embeddings for remaining ACHE models, overlaid on the
    covalent subset of ProteinReactiveDB (blue). ACHE-1 and ACHE-3 are shown
    in the main text (Figure~\ref{fgr:ache-1}).}
\end{figure}

\section{Rediscovered Covalent Inhibitors}
\label{app:rediscovered}

\begin{longtable}{|c|c|c|}
    \caption{Examples of rediscovered EGFR covalent inhibitors.}
    \label{tbl:egfr-rediscovered} \\
    \hline
    \textbf{Model} & \textbf{CovalentInDB ID} & \textbf{Structure} \\
    \hline
    \endfirsthead
    \multicolumn{3}{c}{{\bfseries \tablename\ \thetable{} - continued}} \\
    \hline
    \textbf{Model} & \textbf{CovalentInDB ID} & \textbf{Structure} \\
    \hline
    \endhead
    \hline \multicolumn{3}{r}{{Continued on next page}} \\ \endfoot
    \hline \endlastfoot
    EGFR-1 & CI002163 & \includegraphics[width=0.38\textwidth]{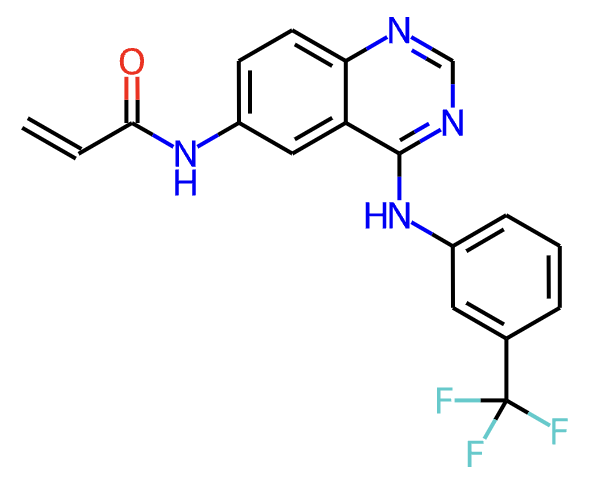} \\
    \hline
    EGFR-1 & CI002725 & \includegraphics[width=0.38\textwidth]{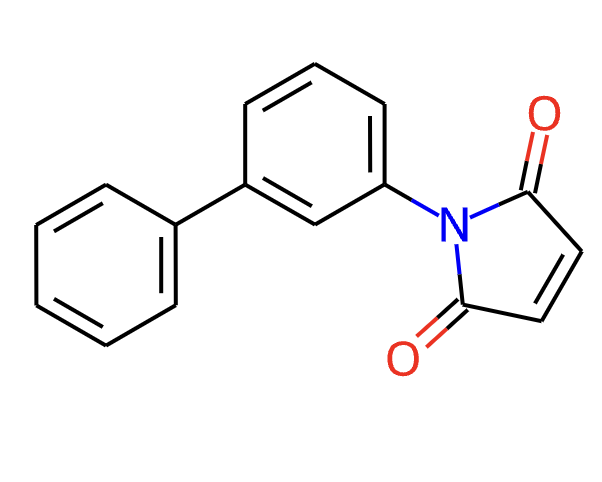} \\
    \hline
    EGFR-2 & CI000048 & \includegraphics[width=0.20\textwidth]{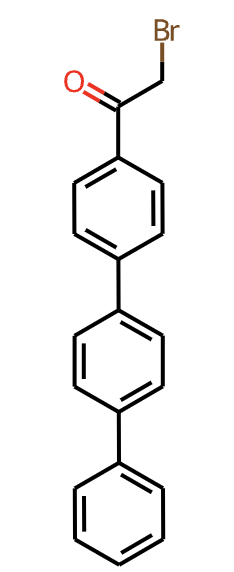} \\
    \hline
    EGFR-3 & CI005895 & \includegraphics[width=0.38\textwidth]{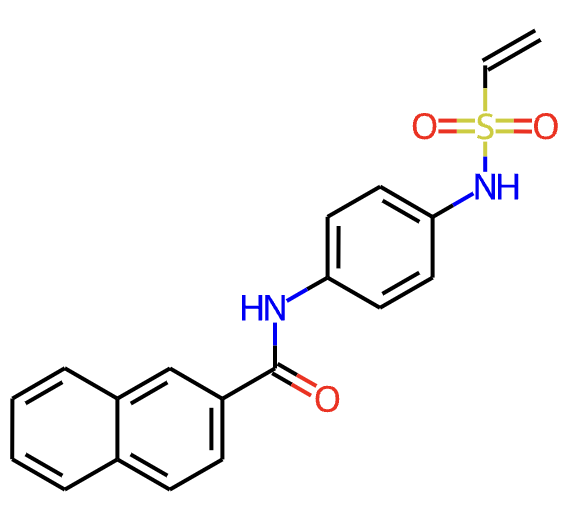} \\
\end{longtable}

\begin{longtable}{|c|c|c|}
    \caption{Examples of rediscovered ACHE covalent inhibitors.}
    \label{tbl:ache-rediscovered} \\
    \hline
    \textbf{Model} & \textbf{CovalentInDB ID} & \textbf{Structure} \\
    \hline
    \endfirsthead
    \multicolumn{3}{c}{{\bfseries \tablename\ \thetable{} - continued}} \\
    \hline
    \textbf{Model} & \textbf{CovalentInDB ID} & \textbf{Structure} \\
    \hline
    \endhead
    \hline \multicolumn{3}{r}{{Continued on next page}} \\ \endfoot
    \hline \endlastfoot
    ACHE-1 & CI005642  & \includegraphics[width=0.38\textwidth]{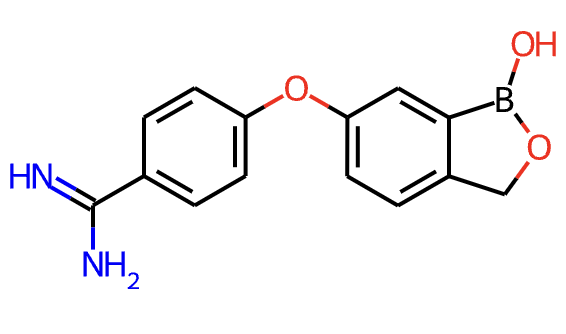} \\
    \hline
    ACHE-1 & CBR002817 & \includegraphics[width=0.28\textwidth]{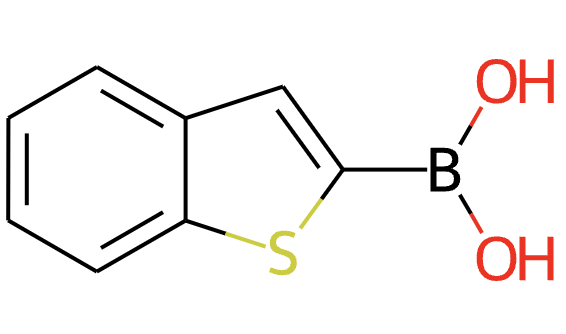} \\
    \hline
    ACHE-3 & CI001308  & \includegraphics[width=0.38\textwidth]{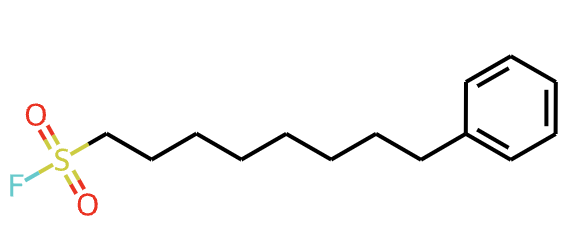} \\
\end{longtable}

\section{Effect of Generation Volume}
\label{app:genvolume}

\begin{table}[H]
    \centering
    \caption{EGFR-3 performance at different generation scales.}
    \label{tbl:more-generated-structures-egfr}
    \begin{tabular}{rrrr}
        \toprule
        \textbf{Generated} & \textbf{Desirable} & \textbf{Rediscovered} & \textbf{Rate (\%)} \\
        \midrule
        1,000    & 245    & 5   & 2.04 \\
        10,000   & 4,793  & 24  & 0.50 \\
        50,000   & 12,930 & 90  & 0.70 \\
        100,000  & 29,381 & 418 & 1.42 \\
        \bottomrule
    \end{tabular}
\end{table}

\begin{table}[H]
    \centering
    \caption{ACHE-3 performance at different generation scales.}
    \label{tbl:generated-structures-ache}
    \begin{tabular}{rrrr}
        \toprule
        \textbf{Generated} & \textbf{Desirable} & \textbf{Rediscovered} & \textbf{Rate (\%)} \\
        \midrule
        1,000    & 26    & 4   & 15.38 \\
        10,000   & 945   & 7   & 0.74 \\
        50,000   & 800   & 78  & 9.75 \\
        100,000  & 1,471 & 113 & 7.68 \\
        \bottomrule
    \end{tabular}
\end{table}

\end{appendices}

\newpage
\bibliographystyle{unsrt}
\bibliography{references}

\end{document}